\documentclass[10pt,twocolumn,letterpaper]{article}

\usepackage{iccv}
\usepackage{times}
\usepackage{epsfig}
\usepackage{graphicx}
\usepackage{amsmath}
\usepackage{amssymb}
\usepackage{gensymb}
\usepackage{ctable}
\usepackage{esvect}
\usepackage{array}
\usepackage{ragged2e}

\newcolumntype{C}[1]{>{\centering\let\newline\\\arraybackslash\hspace{0pt}}m{#1}}

\usepackage[breaklinks=true,bookmarks=false]{hyperref}

\iccvfinalcopy 

\def\iccvPaperID{1566880001704} 

\ificcvfinal\fi

\begin{document}

\title{Street Crossing Aid Using Light-weight CNNs for the Visually Impaired}

\author{Samuel Yu \footnote{} \hspace{2em} Heon Lee \hspace{2em} JungHoon Kim\\
Shanghai American School Puxi West\\
258 Jinfeng Road, Minhang District, Shanghai, China 201107 \\
{\tt\small samuelyu@andrew.cmu.edu, heon\_lee@brown.edu, junghoonkim@snu.ac.kr} 
}

\maketitle
\ificcvfinal\thispagestyle{empty}\fi

\begin{abstract}
   In this paper, we address an issue that the visually impaired commonly face while crossing intersections and propose a solution that takes form as a mobile application. The application utilizes a deep learning convolutional neural network model, LytNetV2, to output necessary information that the visually impaired may lack when without human companions or guide-dogs. A prototype of the application runs on iOS devices of versions 11 or above. It is designed for comprehensiveness, concision, accuracy, and computational efficiency through delivering the two most important pieces of information, pedestrian traffic light color and direction, required to cross the road in real-time. Furthermore, it is specifically aimed to support those facing financial burden as the solution takes the form of a free mobile application. Through the modification and utilization of key principles in MobileNetV3 such as depthwise seperable convolutions and squeeze-excite layers, the deep neural network model achieves a classification accuracy of $96\%$ and average angle error of 6.15\degree, while running at a frame rate of 16.34 frames per second. Additionally, the model is trained as an image classifier, allowing for a faster and more accurate model. The network is able to outperform other methods such as object detection and non-deep learning algorithms in both accuracy and thoroughness. The information is delivered through both auditory signals and vibrations, and it has been tested on seven visually impaired and has received above satisfactory responses. 
\end{abstract}


\section{Introduction}
Modern technology has revolutionized countless fields of study including that of computer vision and artificial intelligence. As neural networks start achieving both a higher accuracy and lower latency, models on mobile devices become more accessible and feasible for real life application. One such field that has been heavily affected is assistive technology for the visually impaired.

A predominant concern that the visually impaired face concerns with information that cannot be attained through white-canes. The most relevant example deals with crossing intersections. This problem is caused by the technical limitations of white-canes: the lack of ability to detect traffic lights or, in general, any feature that can not be attained through physical touch. 

Physical products such as glasses and applications have been developed to target the concern, including those by companies like eSight, Seeing AI, and the Sound of Vision System. 

eSight is a pair of glass that displays everything on two near-to-eye displays that a camera captures in real time \cite{20}. Seeing AI has numerous features such as reading texts, recognizing friends, and describing the surrounding scene \cite{21}. Caraiman et al. \cite{22} proposed the Sound of Vision System, a wearable device that reconstructs the sensed environment and segmentation into objects of interest and conveying the necessary information through auditory and tactile representation. The device is able to identify elements such as signs, texts, doors, and pedestrian crossing. 

There are two notable drawbacks in these solutions however. Firstly, the hardware attempt to take a holistic account of the surrounding so it is not actively looking for pedestrian traffic lights. This means that most of the time, it identifies unnecessary objects around instead of what the visually impaired require, thus rendering as unreliable. Secondly, it poses a financial burden that is especially true for those living in less economic developed countries as the glass costs approximately 10,000 U.S. dollars.

Thus, instead of a hardware solution, this paper discusses a solution taking the form of software. This paper introduces an iOS application that deploys and runs a deep convolutional neural network model locally. 

The neural network, LytNetV2, is trained with two specific goals in mind: providing a comprehensive and feasible solution. It provides a comprehensive solution by outputting the mode of the traffic lights as well as the direction of the zebra crossing in real time. The application takes the output values from the model to inform users when it is safe to cross and whether they are in the correct position and orientation in relation to the zebra-crossing. For the software to render as a feasible solution, it must also be able to run at real-time in a phone.

With both goals in mind, the network is trained as an image classifier as they are both more computationally efficient and accurate compared to object detectors and semantic segmentation. Moreover, to ensure that the speed of the model is able to run on a phone at real-time, the network has been trained on a modified version of Mobilenet V3. In doing so, the network is able to take advantage of features such as depthwise separable convolutions, inverted residuals and linear bottlenecks, and squeeze-excite layers. User-interface has also been carefully considered, and so the application converts the network's outputs into simple audio and tactile information. To further ensure that our application is a viable alternative to current products, it has been tested by the blind for a holistic assessment of reliability, comfort, and simplicity. 

The rest of the paper is organized in the following manner: Section II reviews previous work and contributions made to the development of various products designed to aid the visually impaired; Section III presents the proposed method to achieve the best results for our task; Section IV shows experiments used during and after the creation of the network and compares it against other methods; Section V provides insight to the application and demonstrates experiment results; Section V concludes the paper and examines potential future directions. 

\section{Related Works}
The visually impaired face key navigation issues, one of which includes crossing the road. There have been many attempts to alleviating such privation. One such attempt has been the acoustic pedestrian traffic lights \cite{3, 4, 5}. These specific traffic lights uses audible tones, verbal messages, and/or vibrating surfaces to communicate essential information such as the color of the pedestrian traffic light, and thus, whether it is safe to cross or not \cite{4}. Though acoustic pedestrian traffic lights certainly has its advantages in that the information provided is near-perfectly reliable, its disadvantage lays in the fact that it is currently mostly only implemented in urban areas, and even in urban areas may not be ubiquitous \cite{3}. As a result, the visually impaired cannot solely rely on the presence of an acoustic  pedestrian traffic light while traveling, especially those living in poorer neighborhoods.

The task of detecting traffic light for autonomous driving has been explored by many and has developed over the years \cite{6,7,8,9}. Because the task of detecting traffic light for autonomous driving is closely related to the task of detecting pedestrian traffic light for navigation support, as the technology for autonomous cars improves, so does the technology for crossing streets \cite{25}. Behrendt et al. \cite{10} created a model that is able to detect traffic lights as small as $3 \times 10$ pixels and with relatively high accuracy. Though most models for traffic lights have a high precision and recall rate of nearly 100\% and show practical usage, the same cannot be said for pedestrian traffic lights. Pedestrian traffic lights differ because they are complex shaped and usually differ based on the region in which the pedestrian traffic light is placed. Traffic lights, on the other hand, are simple circles.

Mocanu et al. \cite{23} proposed a solution through the creation of an automatic cognition system that uses computer vision with deep convolution neutral networks to recognize objects. The software aims to mitigate difficulties that the blind face as a whole, none specifically aiming to help with crossing the road, and hence, not being able to provide a feasible solution. 

Shioyama et al. \cite{11} were one of the first to develop an algorithm to detect pedestrian traffic lights and the length of the zebra-crossing. Others such as Mascetti et al. and Charette et al. \cite{3,15} both developed an analytic image processing algorithm, which undergoes candidate extraction, candidate recognition, and candidate classification. Cheng et al. \cite{5} proposed a more robust real-time pedestrian traffic lights detection algorithm, which gets rid of the analytic image processing method and uses candidate extraction and a concise machine learning scheme. 

A limitation that many attempts faced was the speed of hardware. Thus, Ivanchenko et al. \cite{12} created an algorithm specifically for mobile devices with an accelerator to detect pedestrian traffic lights in real time. Angin et al. \cite{13}  incorporated external servers to remove the limitation of hardware and provide more accurate information. Though the external servers are able to run deeper models than phones, it requires fast and stable internet connection at all times. Moreover, the advancement of efficient neural networks such as MobileNets enable a deep-learning approach to be implemented on a mobile device \cite{14, 26}.  

Diaz et al. \cite{24} proposes an intelligent assistive agent to aid the visually impaired to stay within the crosswalk when crossing an intersection. Although, the idea of providing the user with various instructions with live feed is undoubtedly beneficial for the users, the current accuracy, $82\%$, for this model is not optimal for public usage.

Direction is another factor to consider when helping the visually impaired cross the street. Though the visually impaired can have a good sense of the general direction to cross the road in familiar environments, relying on one's memory has its limitations \cite{16}. Therefore, solutions to provide specific direction have also been devised. Other than detecting the color of pedestrian traffic lights, Ivanchenko et al. \cite{16} also created an algorithm for detecting zebra crossings. The system obtains information of how much of the zebra-crossing is visible to help the visually impaired know whether or not they are generally facing in the correct direction, but it does not provide the specific location of the zebra crossing. Poggi et al., Lausser et al., and Banich \cite{17,18,19} also use deep learning neural network within computer vision to detect zebra crossings to help the visually impaired cross streets.

We previously developed LytNet by modifying MobileNetV2 as a deep learning method that outputs both traffic light and zebra crossing information \cite{27}. While the network achieved decent results with an accuracy of 94\%, the task of classifying pedestrian traffic lights demands an even higher accuracy, with the visually-impaired's safety at risk. 

\section{Proposed Method}
With our network required to run at fast speeds on a mobile phone, we adapted our network off of the high-performing baseline of MobileNetV3. 
\subsection{Depthwise Separable Convolutions}
LytNetV2 utilizes depthwise seaprable convolutions, a key aspect of MobileNetV3. Depthwise separable convolution are split in two parts: a "depthwise" convolution and a pointwise convolution (regular convolution of kernel size $1 \times 1$).  In a "depthwise" convolution, the channels of the input image are separated and different filters are used for every convolution over each channel. Then, a pointwise convolution is used to collapse the channels to a depth of 1. For an input of dimensions $h_i \cdot w_i \cdot d_i$ convolved with stride 1 with a kernel of size $k \cdot k$ and $d_j$ output channels, the cost of a standard convolution is $h_i \cdot w_i \cdot k^2 \cdot d_i \cdot d_j$ while the cost of a depthwise separable convolution is $h_i \cdot w_i \cdot d_i \cdot (k^2 + d_j)$ \cite{14}. Thus, the total cost of a depthwise separable convolution is $\frac{k^2 \cdot d_j}{k^2+d_j}$ times less than a standard convolution while being able to maintain a similar level of performance \cite{14}.
\subsection{Squeeze-Excite Layers}
To improve our network, squeeze-excite layers have been used. Squeeze-excite layers improves channel interdependencies at almost no computational cost. With squeeze-excite layers, content aware mechanism is added. In other words, the network places emphasis on more important features while weighing the less important ones less. This is done through adding paramaters to each channel of a convolutional block, allowing for the network to adjust the weighting of each feature adaptively.

\setlength{\arrayrulewidth}{.1em}
\begin{table}
\begin{center}
\begin{tabular}{cc m{0.28cm} c c m{0.3cm} m{0.3cm} m{0.28cm}}
\hline
Input & Operator & k & e & c & SE & NL & s\\
\hline
$768 \times 3$ & conv2d & 3 & - & 16 & - & HS & 2 \\
$384 \times 16$ & maxpool & 2 & - & - & - & - & 2 \\
$384\times 16$ & bneck & 3 & 16 & 16 & - & RE & 1 \\
$192 \times 16$ & bneck & 3 & 64 & 24 & - & RE & 2 \\
$96 \times 24$ & bneck & 3 & 72 & 24 & - & RE & 1 \\
$96 \times 24$ & bneck & 5 & 72 & 40 & \checkmark & RE & 2 \\
$48 \times 40$ & bneck & 5 & 120 & 40 & \checkmark & RE & 1 \\
$48 \times 40$ & bneck & 3 & 240 & 80 & - & HS & 2 \\
$24 \times 80$ & bneck & 3 & 200 & 80 & - & HS & 1 \\
$24 \times 80$ & bneck & 3 & 480 & 112 & \checkmark & HS & 1 \\
$24 \times 112$ & bneck & 5 & 672 & 160 & \checkmark & HS & 2 \\
$12 \times 160$ & bneck & 5 & 960 & 160 & \checkmark & HS & 1 \\
$12 \times 160$ & bneck & 3 & 960 & 320 & - & RE & 1 \\
$12 \times 320$ & conv2d & 1 & - & 960 & - & HS & 1 \\
$12 \times 960$ & avgpool & - & - & - & - & - & - \\
$1^2 \times 960$ & conv2d & 1 & - & 1280 & - & HS & 1 \\
$1280$ & FC & - & - & 5, 4 & - & - & - \\
\hline
\end{tabular}
\end{center}
\caption{Specification for our neural network. The bneck operator denotes the bottleneck block as defined in \cite{26}. k denotes the kernel size. e denotes the expansion size. The FC operator denotes a fully connected layer. c denotes the number of output channels. SE denotes whether Squeeze-And-Excite is used in the block. NL denotes the type of non-linearity. s denotes the stride. The last fully connected layer has two outputs of dimension 5 and 4, representing the number of class predictions and the predicted coordinates respectively. Most inputs are written in the form $w\times c$ to save space, but the real input is $w \times h \times c$, where the height is $\frac{3}{4} \cdot w$.}
\label{table:networkStructure}
\end{table}
\subsection{Training Procedure}
\begin{figure}
    \centering
    \includegraphics[width=\columnwidth]{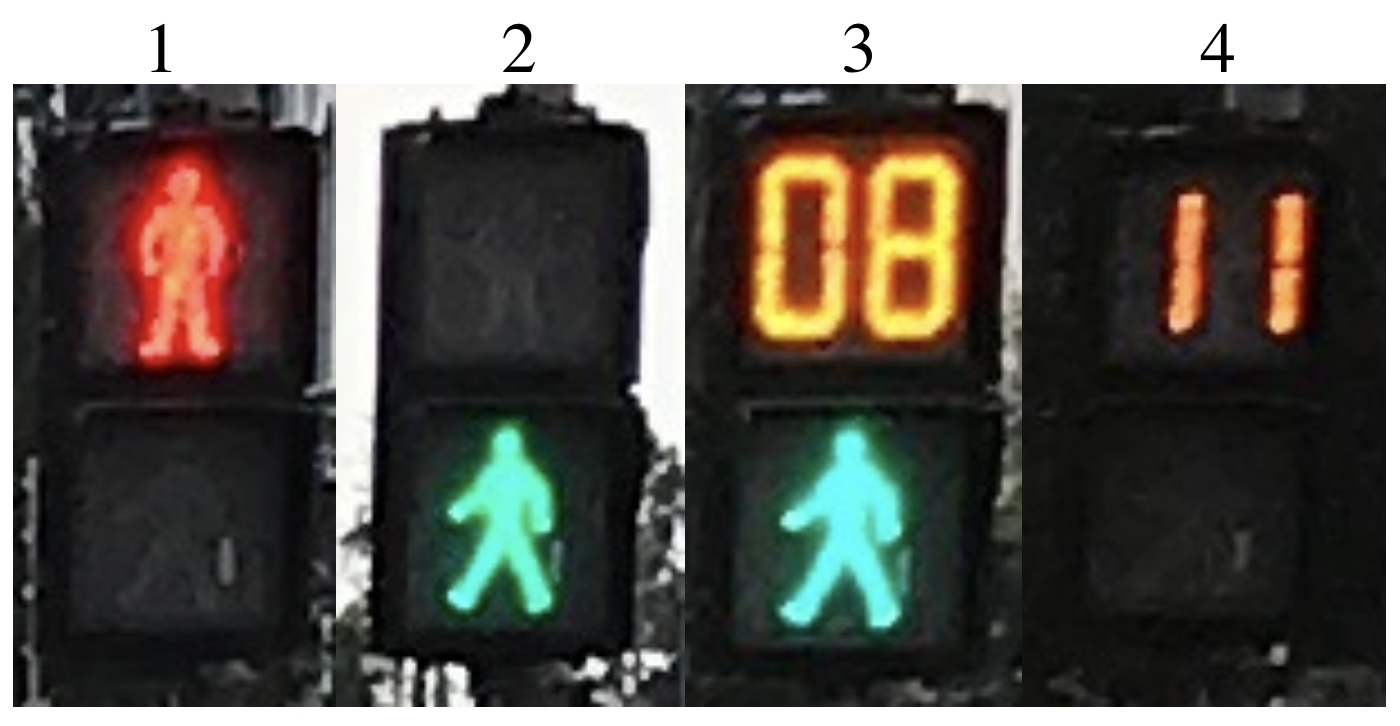}
    \caption{Examples of each class. 1 is red, 2 is green, 3 is countdown green, and 4 is countdown blank. The none class is simply an image without a pedestrian traffic light in it.}
    \label{fig:classSamples}
\end{figure}

Our network was trained using our Pedestrian-Traffic-Light (PTL) Dataset \cite{27}. This dataset is made up of 5059 images, with 3456 images used for training, 864 for validation, and 729 for testing. The images in the dataset have a resolution of $876 \times 657$, labelled with one of five classes: red, green, countdown green, countdown blank, and none. Each image is also labelled with four coordinates: $[x_s,y_s,x_e,y_e]$, where $(x_s, y_s)$ and $(x_e, y_e)$ represent the start and end-point of the mid-line of the zebra crossing in the image. See Figure~\ref{fig:classSamples} for examples of the classes.

\subsubsection{Data Transformations}
With the original images having a resolution of $876 \times 657$, a $768 \times 576$ section was randomly cropped from the original image for each training iteration. The direction vectors were also clipped accordingly. Next, a random horizontal flip of probability 0.5 was applied. 

After training our original network, we analyzed our network's common mistakes. The two most common mistakes were mistaking a countdown blank light with a red light and vice versa. This is because in both red and countdown blank lights, only the top half of the light lights up, so both symbols are in the same position. When the image is of lower quality, the color of both can be very similar, where both are on a spectrum between yellow and red. To help our network differentiate the two classes, we also used a random brightness, saturation, contrast, and hue shift. The idea behind this transformation is that primarily with red and countdown blank samples, the color of the light will occasionally be shifted from red to yellow or vice versa, forcing the network to put more weight in learning the shape of the red man or yellow number(s) have less of a reliance on the color. 
\subsubsection{Loss Function}
With our network designed to do two different tasks, one which is classification and the other which is regression, our loss function combines classification and regression loss as done in \cite{27}. Our classification loss is cross-entropy loss, defined as:
\begin{equation}
    L_{cls}(t,p) = -\sum_{c=1}^{N} y(c,t) \log p(c),
\end{equation}
where $t$ is the ground truth class, $p$ is the prediction, $y(c,t)$ is a binary indicator if $c=t$, and $p(c)$ is the probability of class $c$. Our regression loss is mean-squared-error loss, defined as:
\begin{equation}
    L_{reg}(t,p) = \frac{1}{n}\sum_{i=1}^{n}(t_i - p_i)^2,
\end{equation}
where $t$ is the ground truth and $p$ is the predicted array of coordinates. Combining the two losses, we define our loss function as follows:
\begin{equation}
    L(t,p) = \lambda \cdot L_{cls}(t,p) + (1-\lambda) \cdot L_{reg}(t,p),
\end{equation}
where lambda is a tunable hyper-parameter. We trained our network with the value $\lambda = 0.4$, which we determined was the value that provided an optimal balance between the two losses, and provided for the best training results. Our network was trained on a single RTX 2080ti with a batch size of 32. We used the Adam optimizer with an initial learning rate of 0.001 and learning rate drops of a factor of 10 at epochs: $[400, 700, 1000,1300]$, and full convergence at 1600 epochs.

As we have done in \cite{27}, some special evaluation metrics are used to gain a better understanding of the networks performance. With regards to the zebra crossing midline prediction, we use three metrics, angle-error, startpoint-error, and endpoint-error. Given the predicted coordinates $(x_p, y_p)$ and the ground truth $(x_t, y_t)$, the startpoint and endpoint error is simply the distance formula.

To calculate the angle between the two midline predictions, we use the dot product of vectors. We convert the predicted coordinates $(\hat{x}_s, \hat{y}_s)$  and $(\hat{x}_e, \hat{y}_e)$, we establish the predicted direction vector 
\begin{equation}
    \vv{p} = \left(\hat{x}_e - \hat{x}_s, \hat{y}_e - \hat{y}_s\right).
\end{equation} 
The same is done with the ground truth coordinates to create the ground truth direction vector $\vv{t}$. Thus, we can calculate the angle error $\theta$ with:
\begin{equation}
    \theta = \arccos{\frac{\vv{p} \cdot \vv{t}}{\lvert \vv{p} \rvert \cdot \lvert \vv{t} \rvert}}.
\end{equation}

\section{Experiments} 
\subsection{Ablation Study}
Starting from MobileNetV3, we made many changes to the network to optimize it for our task of pedestrian traffic light and zebra crossing detection. 
\subsubsection{MaxPooling After First Conv}
With the baseline MobileNetV3 designed to run on inputs of size $224 \times 224 \times 3$, changes must be made to the network to allow for the network to run at acceptable frame rates on mobile phones using an input of size $768 \times 576 \times 3$. In theory, the use of a $2 \times 2$ MaxPool reduces the input to the main body of the network by a factor of 4, greatly reducing the required computational time, despite not entirely catching up to the speeds of stock MobileNetV3 on a $224 \times 224 \times 3$ input. 

As seen in Table~\ref{table:maxpoolComparison}, the use of a MaxPool does result in a slight decrease in the network's performance. This is to be expected, as a MaxPool layer results in less information available to be input through the middle of the network. However, the benefit comes in the 34\% decrease in inference time. This allows our network to approach the necessary speed to be able to run close to real-time. 
\begin{table}
\begin{center}
\begin{tabular}{cccccc}
\hline
Network & Accuracy & a-err & sp-err & ep-err & t\\
\hline
Baseline & \textbf{94.45} & \textbf{7.63} & \textbf{0.094} & \textbf{0.056} & 10.97\\
MaxPool & 94.05 & 8.04 & 0.099 & 0.063 & \textbf{7.28}\\
$\Delta$ & -.4 & +.41 & +.005 & +.007 & -3.69\\
\hline
\end{tabular}
\end{center}
\caption{Comparison between baseline MobileNetV3 network and the addition of a 2x2 maxpool layer. a-err denotes average angle error. sp-err denotes average startpoint error. ep-error denotes average endpoint error. t denotes inference time in seconds required for 729 testing images on a GTX 1080.}
\label{table:maxpoolComparison}
\end{table}

\subsubsection{Pruning Unnecessary Layers}
\begin{table}
\begin{center}
\begin{tabular}{cccccc}
\hline
Network & Accuracy & a-err & sp-err & ep-err & t\\
\hline
MaxPool & 94.05 & 8.04 & 0.099 & 0.063 & 7.28\\
Pruned & \textbf{94.68 }& \textbf{6.94} & \textbf{0.092} & \textbf{0.059} & \textbf{6.88}\\
$\Delta$ & +.63 & -1.1 & -.007 & -.004 & -.4\\
\hline
\end{tabular}
\end{center}
\caption{Comparison between network with MaxPooling and network with less layers.}
\label{table:pruneLayerComparison}
\end{table}
For the simpler task of classifying pedestrian traffic lights, we hypothesize that it is not necessary to use all the stride=1 layers in MobileNetV3, which was designed for the more challenging classification task on ImageNet. By only removing some stride=1 layers, the basic network structure is still maintained, while the speed of the network will increase. We removed 5 layers from the original MobileNetV3 structure. The layers were removed such the the network maintained the structure of the number of channels (or feature maps) throughout the network, while having at most 2 layers with stride=1 following each stride=2 layer. This allows for the structural increase in the depth of the network to stay the same. 

Table~\ref{table:pruneLayerComparison} shows that our hypothesis was correct, with the network not only decreasing in inference time, but also improving in performance. We believe that the extra stride=1 layers were helping the network over-fit; thus, removing them made convergence more difficult but improved validation accuracy.
\subsubsection{Adding More Depth to the Network}
\begin{table}
\begin{center}
\begin{tabular}{cccccc}
\hline
Network & Accuracy & a-err & sp-err & ep-err & t\\
\hline
Pruned & 94.68 & \textbf{6.94} & \textbf{0.092} & \textbf{0.059} & 6.88\\
Deeper & \textbf{95.94} & 6.15 & 0.076 & 0.048 & \textbf{7.12}\\
$\Delta$ & +1.26 & -.79 & -.016 & -.011 & +.24\\
\hline
\end{tabular}
\end{center}
\caption{Comparison between pruned network and network with an extra layer outputting 320 channels.}
\label{table:depthComparison}
\end{table}
In MobileNetV3, the 160 channels output from the last bottleneck block is directly expanded to 960 channels using a $1\times1$ conv2d operation. This was designed as a more efficient last stage, replacing the original last stage which output 320 channels after the last bottleneck block \cite{26}. While there was no performance difference in \cite{26}, we nevertheless tested using 320 channels after the last bottleneck block, which was shown to be effective in MobileNetV2 and LytNet \cite{14, 27}. With doubling the number of channels, we saw a marked increase in all aspects of performance, while coming with a small computational cost (Table~\ref{table:depthComparison}). 
\subsection{Comparison With Other Methods}
We first compare our network, LytNetV2, to LytNet on the PTL Dataset. As shown in Table~\ref{table:PTLAccuracyErrorComparison}, our network outperforms LytNet, which was based on MobileNetV2, in every metric, but it comes at the cost of extra inference time. The increase in network accuracy is paramount, because the visually impaired must receive accurate information to safely cross the road. Thus, we argue that the increase in accuracy outweighs the increase in inference time, because even with a 18\% increase, the network can still run at a decent speed ($>15$ fps) on mobile devices, making this tradeoff very worthwhile. Furthermore, LytNet was tested using larger width multipliers which increases the complexity of the network, and was still unable to achieve higher performance \cite{27}. Our network breaks through this bottleneck and is able to come closer to the ultimate goal of this task, which is to provide accurate information to the visually impaired while being able to deliver such information in a convenient manner, and taking an acceptable amount of time. 

\begin{table}
\begin{center}
\begin{tabular}{cccccc}
\hline
Network & Accuracy & a-err & sp-err & ep-err & t\\
\hline
LytNet & 94.18 & 6.27 & 0.076 & 0.051 & \textbf{6.05}\\
LytNetV2 & \textbf{95.94} & \textbf{6.15} & 0.076 & \textbf{0.048} & 7.12\\
$\Delta$ & +1.76 & -.12 & -.000 & -.003 & +1.07\\
\hline
\end{tabular}
\end{center}
\caption{Comparison between LytNetV2 and LytNet.}
\label{table:PTLAccuracyErrorComparison}
\end{table}

Further analysis shows that our network is able to outperform LytNet in precision and accuracy in almost every single class, with only the precision for the None class being higher on LytNet. Our network is able to greatly increase in the weak performance of LytNet on the countdown blank class, thus removing a large weakness of LytNet. (Table~\ref{table:PTLPrecisionAccuracyComparison}). 

\begin{table}
\begin{center}
\begin{tabular}{p{1.25cm} p{1.1cm} p{0.6cm} p{0.6cm} p{0.6cm} p{0.6cm} p{0.6cm}}
\hline
& Network & Red & Green & CDG & CDB & None\\
\hline
Precision & Ours & \textbf{0.98} & \textbf{0.95} & 0.99 & \textbf{0.93} & 0.91\\
& LytNet & 0.97 & 0.94 & 0.99 & 0.86 &\textbf{0.92}\\
Recall & Ours & 0.96 & \textbf{0.96} &\textbf{0.97} & \textbf{0.97} & \textbf{0.91}\\
& LytNet & 0.96 & 0.94 & 0.96 & 0.89 & 0.89\\
F1 Score & Ours & \textbf{0.97}& \textbf{0.95} & \textbf{0.98} & \textbf{0.95}& \textbf{0.91}\\
& LytNet & 0.96 & 0.94 & 0.97 & 0.89 & 0.89\\
\hline
\end{tabular}
\end{center}
\caption{Comparison of precision, accuracy, and F1-score between our network and LytNet on the PTL dataset over all five classes. CDG denotes the countdown green class, and CDB denotes the countdown blank class.}
\label{table:PTLPrecisionAccuracyComparison}
\end{table}

Without even re-training our network, we tested our network on the China portion of the PTLR Dataset, which uses input images of size $1280\times720$ \cite{5}. The only processing we did was to change all network predictions of class "countdown green" and "countdown blank" to be a prediction of "none", because the images in the China section of the PTLR Dataset were only from three different classes: red, green, and none. After testing our network on the dataset, we compared our results to the results from \cite{27} and \cite{5}.

\begin{table}
\begin{center}
\begin{tabular}{ccccc}
\hline
&  & Ours & LytNet & Cheng et al. \\
\hline
Red & Precision & \textbf{98.56}& 96.24 & 96.67\\
& Recall & \textbf{94.36} & 92.23 & 86.43\\
& F1 Score & \textbf{96.41} & 94.19 & 91.26\\
Green & Precision & 97.03 & \textbf{98.83}& 98.03\\
& Recall & \textbf{94.97} & 92.15 & 91.30\\
& F1 Score & \textbf{95.99} & 95.37 & 94.55\\
\hline
\end{tabular}
\end{center}
\caption{Comparison of precision, accuracy, and F1-score between our network and other methods on the PTLR dataset.}
\label{table:PTLRComparison}
\end{table}

As shown in Table~\ref{table:PTLRComparison}, our network outperforms both methods in F1 Score. While we concede that Cheng et al.'s algorithm is still easier adapt to different traffic lights from different countries, we can see that our method is viable in detecting traffic lights with varying features. This is without even adjusting our network by removing the direction prediction and retraining our network on only three classes, as we did in LytNet \cite{27}. Neither did we retrain our network on the PTLR Dataset (primarily due to insufficient data). We argue that the significant increase in performance when comparing our method to Cheng et al.'s justifies the drawback of neural networks, which is the large amount of training data needed to adapt the network. A method must be highly accurate before it can be employed, or else there is a safety risk. Furthermore, our method provides both the traffic light and zebra crossing information at the same time, a significant advantage over Cheng et al.'s algorithm.

\subsection{Sample Incorrect Predictions}
\begin{figure}
    \centering
    \includegraphics[width=\columnwidth]{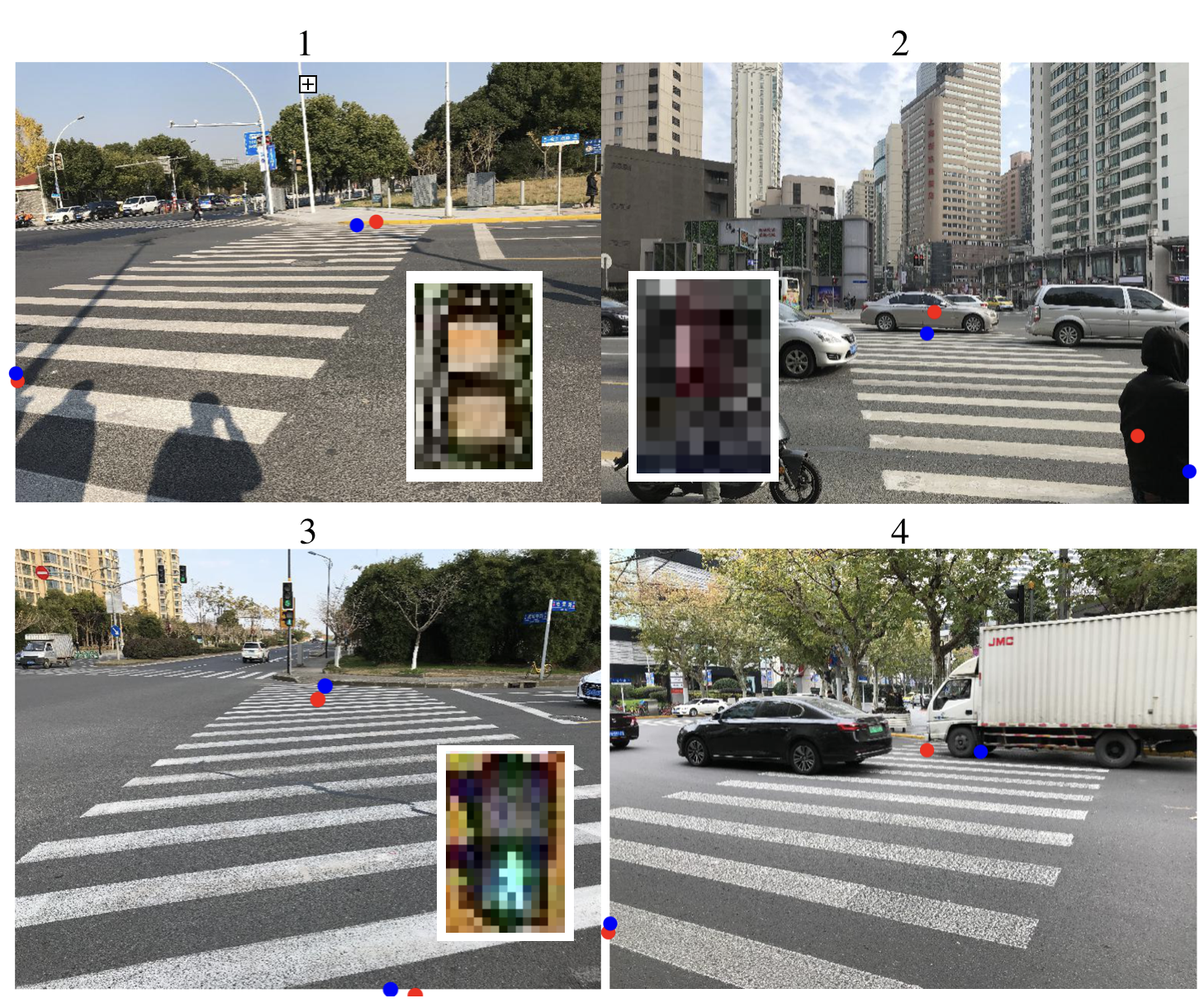}
    \caption{Sample incorrect predictions from our network. The blue dots represent the ground truth endpoints of the zebra crossing's midline, and the red dots represent our network's predictions.}
    \label{fig:incorrectPredictions}
\end{figure}
To get a better idea of the strengths and weaknesses of our network, we analyzed some of our network's predictions, as shown in Figure~\ref{fig:incorrectPredictions}.

In Image 1, the ground truth is red, but the prediction was of countdown blank. The pedestrian traffic light in this image is severely underexposed, with a barely discernible color and no discernible shape. In this case, given the input resolution of $768 \times 576$, it is almost impossible for the network to predict the class correctly, with a prediction of countdown blank even being more realistic in this example because the color of the traffic light is closer to orange-yellow rather than red. In image 2, the ground truth is again red, but the prediction was countdown blank. In this case, the main problem most likely came during the down-sampling of the image from the original $4032 \times 3024$ to $768 \times 576$. Much of the detail of the red person in the traffic light was lost, and what remained was a vertical strip similar to the number 1. Again, predicting the correct class for this example is almost impossible, and it could almost be argued that countdown blank is the more correct class if we don't see that there are moving cars blocking the zebra crossing, signifying a red light. We see that a relatively common mistake for our network is with underexposed traffic lights. To remedy the issues found in the first two images, a higher input resolution must be used.

Image 3 shows a rather surprising mistake by our network. The ground truth is green, but the network predicted countdown green instead. In this case, the network predicted countdown green with a confidence of 0.506, which is a low confidence given we only have five classes. Image 4 shows a similar mistake, where the network surprisingly predicted a red pedestrian traffic light when there was nothing in the image. Again, the confidence in this prediction was only 0.521, while the confidence that there was nothing was 0.479. This brings about another category of mistakes, which are low confidence predictions. To deal with this problem, we add a confidence threshold to our mobile application, as detailed in the next section, ensuring that all predictions are of high confidence. 
\section{Mobile Application}
\begin{figure*}
    \centering
    \includegraphics[width=\textwidth]{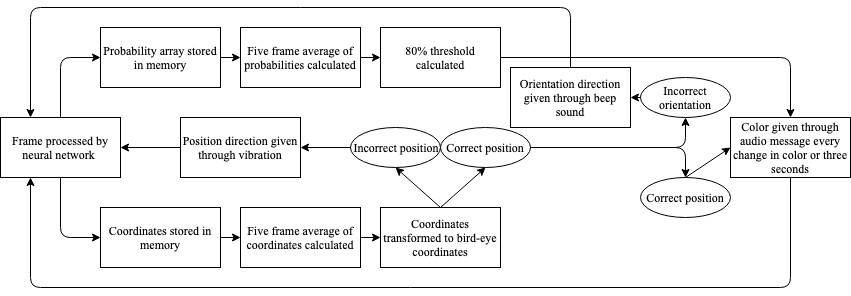}
    \caption{Our application continuously iterates through this flowchart at 16.34 frames per second.}
    \label{fig:flowchart}
\end{figure*}
As a proof of concept, we developed a mobile application that is able to deploy the neural network and output the necessary information in simple ways through the utilization of auditory and tactile methods.
\subsection{Coordinate Conversion}
The model predicts the endpoints of the midpoints of the zebra-crossings. Though this is assumed to be correct, this is only true within a 2-dimensional representation of the world. The appearance of the zebra-crossing in the image plane is an incorrect representation of the position of objects in the 3-dimensional world. Hence, as performed in \cite{27}, for the application to output the correct coordinates in the 3-dimensional world, the image is converted using multiple-view geometry. Since the zebra-crossing is on the ground, the z-value is assumed to be fixed at $z=1$, which enables for the conversion of the image to a 2-dimensional bird-eyes perspective image. To do so, a transformation matrix that is assumed to be consistent is applied to the image. While the height and angle of the image does impact the accuracy of the converted coordinates, for the purpose of helping position and orient the visually impaired, the estimate is sufficient.

\subsection{Application Logic}
To effectively help the visually impaired cross the street, our information is divided into three different stages. The first stage is to correctly position the visually impaired at the midline of the zebra crossing. For positioning, in an image with width $w$ and midline at $(w-1)/2$, if $x_{int} > (w-1)/2 + w\cdot0.085$, users are told to move left, and if $x_{int} < (w-1)/2 - w\cdot0.085$, users are told to move right. Since the edges of the zebra crossing are within $8.5\%$ of the midline in our image, assuming a constant width for the zebra crossing, if $x_{int}$ is outside of the $8.5\%$ range, the user will be outside of the zebra crossing. Understanding that there can be an overload of information if the vibrations happen every second or even faster, our application only re-notifies the visually impaired on whether they should keep moving left or right every two seconds. However, if the instruction changes from moving left to moving right, or vice-versa, the application immediately notifies the user. Furthermore, as soon as the start-point is within the range, a voice message alerts the user.

After correctly positioning the user, our application then orients the user to cross the street in the correct direction. In terms of orientation, a range of $10\degree$ was set for $\Delta\theta$. When $\Delta\theta < -10\degree$, users are directed to rotate left, while when $ \Delta\theta > 10\degree $, users are directed to rotate right. The information of the direction for rotations are delivered through beeps, with a single beeps notifying the user to turn to the left and two quick beeps notifying the user to turn to the right. Similarly to the position, we avoid overloading the visually impaired with information by following the same procedure; we have a two second delay. As soon as the orientation is within the range, we notify the user through a voice message.

Though the network is able to attain a pedestrian traffic light mode recognition of 96\%, the application takes further steps mentioned in \cite{27} to alleviate the risk of detecting the wrong mode of light. First, the softmax probabilities of each class is stored and averaged over five consecutive frames, with countdown\_blank and countdown\_green considered to represent the same mode. Next, a confidence threshold of 0.8 is set, which prevents decisions being made before or after the light changes color as the probability threshold will not be met if one or more frames are of different modes. Finally, after reaching the probability threshold and the user having the correct position and orientation, the application informs the user of the traffic light color through an audio message. If at any time the traffic light mode changes, the application will immediately notify the user. However, if the traffic light mode remains the same, the application will only notify the user once every three seconds. 

The structure of the application is shown in Figure~\ref{fig:flowchart}. 

\subsection{Usability Study}
\begin{figure}
    \centering
    \includegraphics[width=\columnwidth]{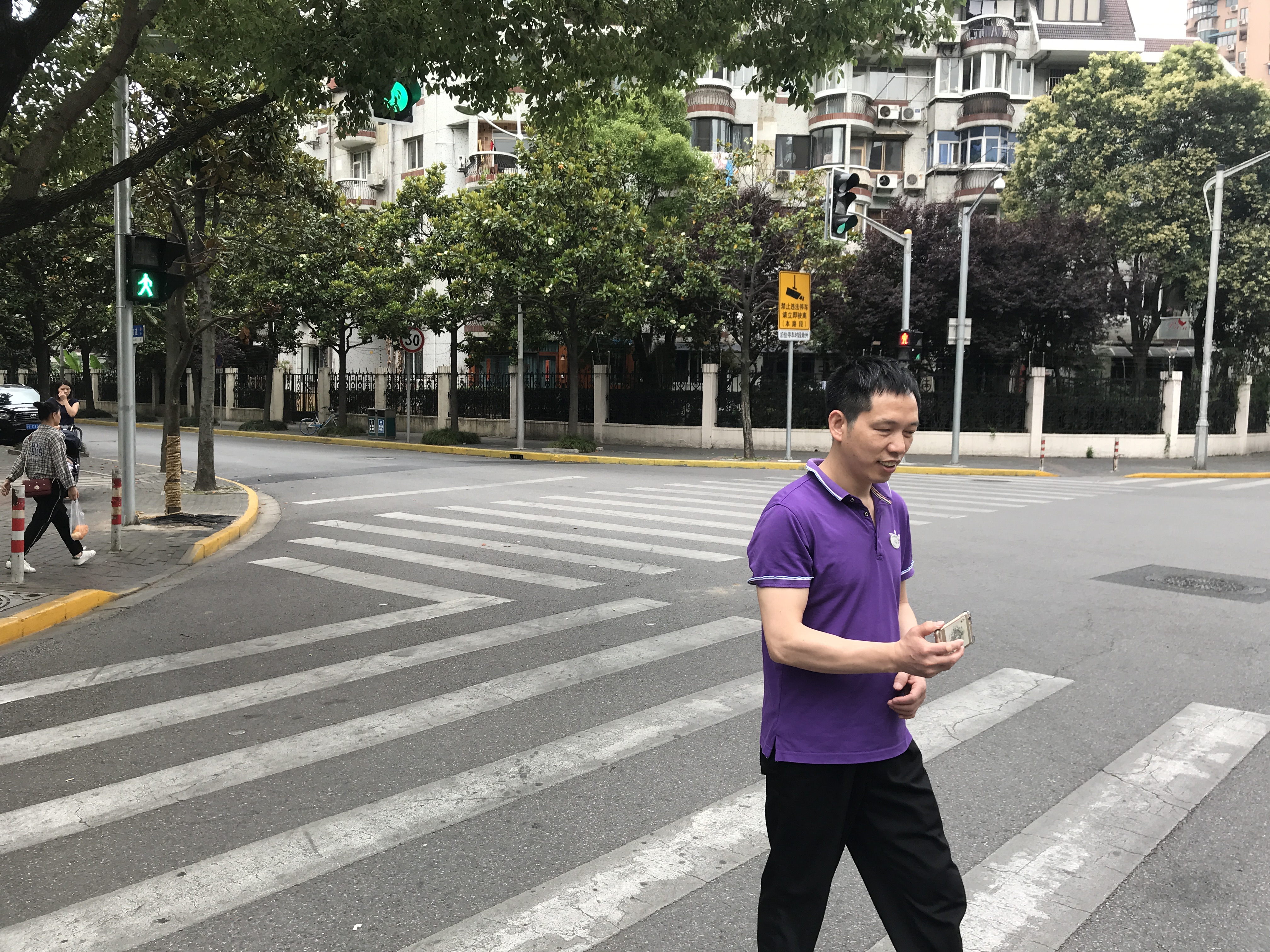}
    \caption{The user holds the phone smartphones horizontally and perpendicular to the ground.}
    \label{fig:person}
\end{figure}
\begin{figure}
    \centering
    \includegraphics[width=\columnwidth]{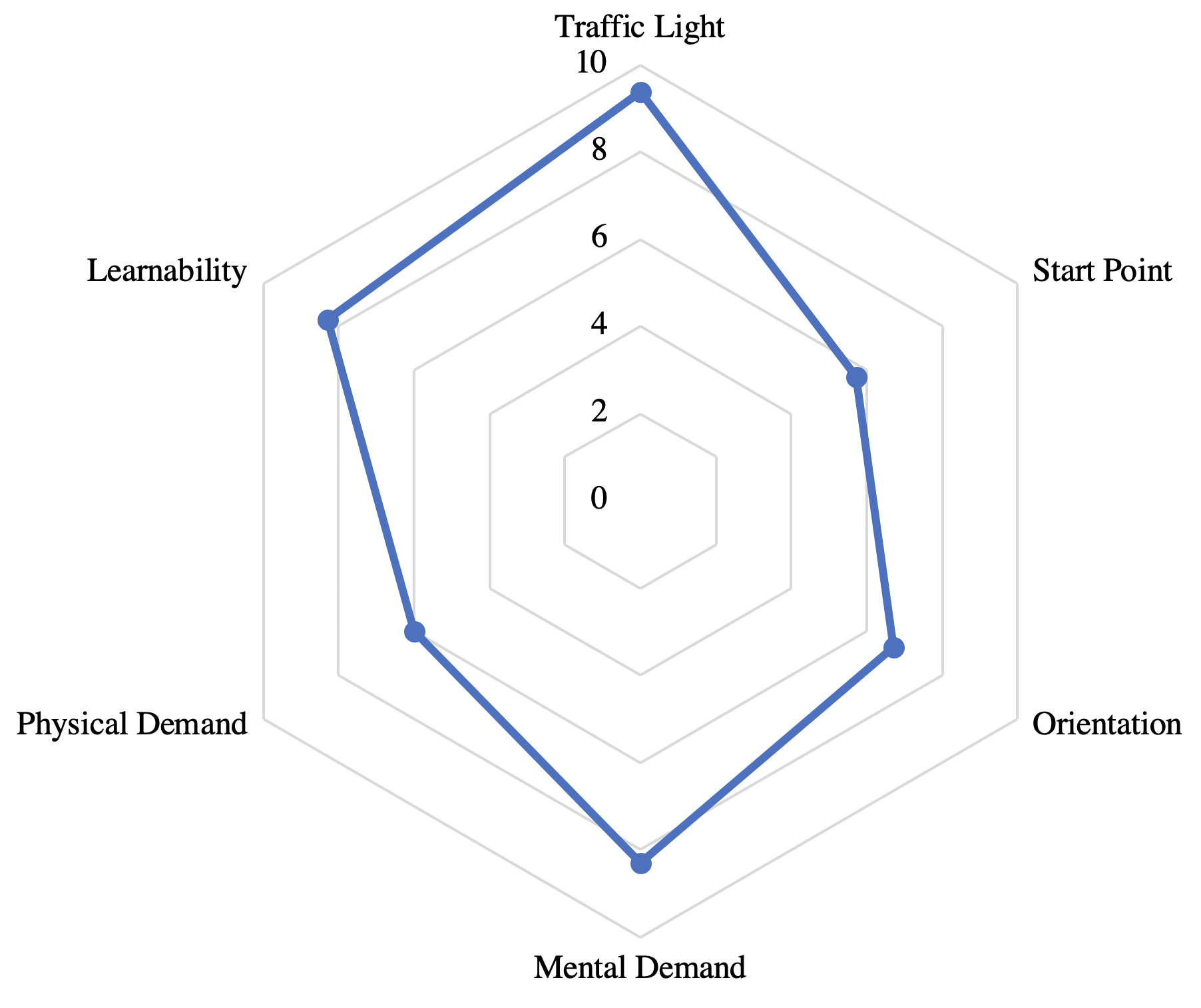}
    \caption{Scores given by blind test subjects for various categories.}
    \label{fig:fig5}
\end{figure}

To evaluate the efficacy of the application, it has been tested on seven visually impaired. As seen in Figure~\ref{fig:person}, the visually impaired is instructed to hold their smartphones horizontally and perpendicular to the ground. After the testing, we collected their feedback based on six categories: traffic light feature, starting point feature, direction feature, learnability, physical demand, and mental demand, asking them to score each category from 1 to 10, with 1 being something that distracts more than it helps, 5 being average, 7 being something they would use in their day-to-day lives, and 10 being something that is perfect and needs no improvements. The radar graph of each category assessing multiple aspects of the application can be seen in Figure~\ref{fig:fig5}. 

For the traffic light feature, the testers found it to be effective and vastly useful, scoring an average of 9.4 out of 10. When asking for comments, there were immediate positive comments about this feature. They found the direction and starting point features to be helpful, but harder to use compared to the traffic light notification, rating the categories with scores of 5.7 and 6.7 respectively. We noticed that it was less intuitive for the visually impaired to respond to the vibrations and beeps compared to outright being told the color of the traffic light. However, they believe that with more time to practice using the application, they will be more accustomed and fluent.

Overall, they rated the learnability highly, with a score of 8.3, citing that after a few minutes of explanation from us, they understood how the app worked. After a couple of guided crossing of the street, the subjects whom travel alone regularly were able to confidently use our application alone. They also rated mental demand with a score of 8.3. At first, the vibrations and beeps may not be completely intuitive, causing the subjects to need some time to think, but after they quickly became accustomed to the application, there was much less mental demand, if any. The physical demand of the device was rated lower, with a score of 6. There was slight annoyances in keeping the phone held in the correct orientation, and also requires both of their hands to be full when walking with a white cane. This small issue can easily be resolved by making a lanyard to hand the phone from their necks, or creating a product to attach the phone onto their body.

\section{Conclusion}
In this paper, we have developed an improved convolutional neural network version of LytNet based off of MobileNetV3 that is able to outperform existing methods. We have shown that our method is robust by testing it on a new dataset without retraining our model, thus countering one of the main drawbacks of neural networks, which is the large amount of data needed to effectively train one. 

Furthermore, we have fully developed a working mobile application that proves our concept is feasible and effective for the visually impaired to use. With three different outputs and instructions required to help the visually impaired cross the street, effectively conveying such information can be challenging, but our tests on the visually impaired show that our method is a good foundation to build upon. 

Further improvements can be made to our network in two different ways. First, images taken of street intersections at night can be added to the dataset. Otherwise, currently the network and application can only be used during the day. Second, additional data from other countries can be used to allow our network to completely generalize among all countries' traffic lights and crosswalks, preventing it from being region-locked. With regards to our application, a simple detector can be added to inform the visually impaired if their finger may be accidentally covering the camera, which we found to be an occasional problem when testing the application. 

{\small
\bibliographystyle{ieee_fullname}
\bibliography{egbib}
}

\end{document}